\documentclass[sn-mathphys-num]{sn-jnl}

\usepackage{graphicx}%
\usepackage{multirow}%
\usepackage{makecell}%
\usepackage{amsmath,amssymb,amsfonts}%
\usepackage{amsthm}%
\usepackage{mathrsfs}%
\usepackage[title]{appendix}%
\usepackage{xcolor}%
\usepackage{textcomp}%
\usepackage{manyfoot}%
\usepackage{booktabs}%
\usepackage{algorithm}%
\usepackage{algorithmicx}%
\usepackage{algpseudocode}%
\usepackage{listings}%
\usepackage{xspace}%

\theoremstyle{thmstyleone}%
\theoremstyle{thmstyletwo}%

\theoremstyle{thmstylethree}%

\newcommand{\etal}{\textit{et al.\xspace}}
\newcommand{\removed}[1]{}

\begin{document}

\title[Uncovering Temporal Patterns in Visualizations of High-Dimensional Data]{Uncovering Temporal Patterns in Visualizations of High-Dimensional Data}

\author*[1]{\fnm{Pavlin G.} \sur{Poli\v{c}ar}}\email{pavlin.policar@fri.uni-lj.si}

\author[1,2]{\fnm{Bla\v{z}} \sur{Zupan}}

\affil*[1]{\orgdiv{Faculty of Computer and Information Science}, \orgname{University of Ljubljana}, \orgaddress{\street{Ve\v{c}na Pot 113}, \city{Ljubljana}, \postcode{1000}, \country{Slovenia}}}

\affil[2]{\orgdiv{Department of Education, Innovation and Technology}, \orgname{Baylor College of Medicine}, \orgaddress{\street{1 Baylor Plz}, \city{Houston}, \postcode{77030}, \state{Texas}, \country{United States}}}

\abstract{
  With the increasing availability of high-dimensional data, analysts often rely on exploratory data analysis to understand complex data sets. A key approach to exploring such data is dimensionality reduction, which embeds high-dimensional data in two dimensions to enable visual exploration. However, popular embedding techniques, such as t-SNE and UMAP, typically assume that data points are independent. When this assumption is violated, as in time-series data, the resulting visualizations may fail to reveal important temporal patterns and trends. To address this, we propose a formal extension to existing dimensionality reduction methods that incorporates two temporal loss terms that explicitly highlight temporal progression in the embedded visualizations. Through a series of experiments on both synthetic and real-world datasets, we demonstrate that our approach effectively uncovers temporal patterns and improves the interpretability of the visualizations. Furthermore, the method improves temporal coherence while preserving the fidelity of the embeddings, providing a robust tool for dynamic data analysis.
}


\maketitle

\section{Introduction}

In recent decades, high-dimensional data have become ubiquitous in various fields, including bioinformatics~\cite{LaManno2018,Kobak2019,Policar2021}, medicine~\cite{Benito2024,Policar2023b}, social sciences~\cite{DaSilva2018,DaSilvaLopes2020}, and engineering~\cite{Li2019}.
To analyze such data, practitioners often rely on visualization techniques to detect patterns and identify relationships between data elements and variables.
For example, histograms can reveal the distribution of a particular variable, while scatterplot matrices can identify correlations between variables. However, these methods are limited to working with a relatively small number of variables at a time, and they usually fail to uncover relationships and interactions that involve a larger group of variables.
When the number of variables is very large, manually examining each variable or pair of variables becomes impractical and time-consuming.
In such cases, analysts often rely on two-dimensional data maps to explore and visualize high-dimensional data.
These maps are created using dimensionality reduction techniques that embed high-dimensional data sets in two dimensions, allowing them to be visualized in scatterplots.

The goal of dimensionality reduction is to obtain a low-dimensional layout that preserves important properties of the original high-dimensional space, such as pairwise distances between data points or the proximity of similar data items in neighborhoods.
Over the years, numerous dimensionality reduction methods have been developed, ranging from linear techniques such as principal component analysis (PCA)~\cite{Jolliffe2002}, linear discriminant analysis (LDA)~\cite{Fisher1936}, and FreeViz~\cite{Demsar2007}, to nonlinear methods such as multidimensional scaling (MDS)~\cite{Kruskal1978}, t-distributed stochastic neighbor embedding (t-SNE)~\cite{Maaten2008}, and uniform manifold approximation and projection (UMAP)~\cite{McInnes2018}.

While dimensionality reduction techniques have been very successful in simplifying and visualizing complex, high-dimensional data sets, they often fail to account for additional relational information between data elements. This includes temporal dependencies or any structural connections between data samples. The standard methods mentioned above treat data as independent collections of points, ignoring any additional information about their relationships. For example, time-series data sets, with observations at points in time that may include measurements of a number of features, inherently contain sequential dependencies that can provide valuable insights into trends, transitions, or periodic behavior. Ignoring these temporal aspects can result in embeddings that misrepresent or obscure the underlying dynamics~\cite{Crnovrsanin2009,Jackle2016}.

\begin{figure}
  \centering
  \includegraphics{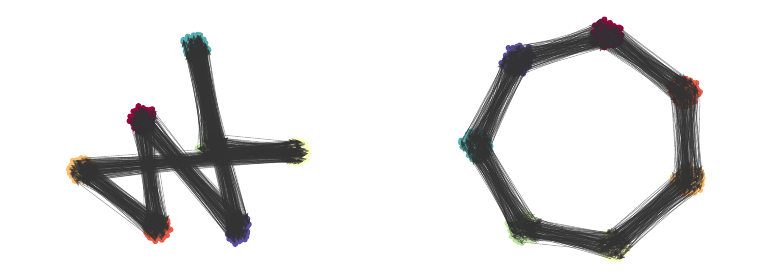}
  \label{fig:starting-example}
  \caption{{\bf A simple example of a two-dimensional embedding of a cyclic data set with arrows connecting consecutive data points.} The panel on the left shows a visualization constructed by embedding the data first and then adding arrows to connect consecutive time-points, while the panel on the right shows the visualization constructed by embedding the data with our proposed embedding that jointly optimizes point positions with respect to the data and temporal information. The cyclic structure is clearly visible in the depiction on the right, and not discernible in the depiction on the left panel.}
\end{figure}

Recently, several approaches have been proposed to integrate dimensionality reduction techniques with sequence information~\cite{Rauber2016,Elzen2016,Ali2018}. In the simplest approach, the data is first transformed into a low, typically two-dimensional space, and the resulting visualization is overlain with arrows connecting related, temporally consecutive data points. However, separating the embedding process from the temporal annotation often results in cluttered visualizations with overlapping arrows pointing in different directions, making it difficult to identify temporal patterns (Fig.~\ref{fig:starting-example}). More advanced methods extend two-dimensional embeddings to a third, temporal dimension. We review these and other related approaches in the Related Work section, emphasizing that none of these methods directly incorporate temporal information into the construction of two-dimensional embeddings.

Here, we propose a direct approach to two-dimensional time-dependent embedding. We present a method that integrates temporal information directly into the embedding construction process. Specifically, we introduce two new loss functions: the directional coherence loss, which favors layouts with non-overlapping arrows pointing in the same direction when positioned close to each other, and the edge length loss, which prefers shorter arrows to produce more coherent visual representations. These losses are incorporated into the t-SNE algorithm, and we demonstrate the method's application to various synthetic and real-world datasets.

We begin with a review of related work in the field, focusing on approaches designed to construct embedded representations with temporal information. We then introduce the notation and the t-SNE algorithm, and present our newly proposed loss functions. We then discuss the optimization procedure and the datasets used in the evaluation. Much of our work involves a critical evaluation of the capabilities of the proposed technique, where we study both the utility of the methods and the impact of various parameters, such as loss function weights, on real and synthetic datasets. We also propose a set of criteria for the quantitative evaluation of the resulting visualizations. The most encouraging result of the evaluation is the interpretability of the constructed embeddings: the visual representations generated by the proposed method are more coherent and visually clear, revealing cycles and other temporal patterns that would otherwise be difficult or impossible to discern using standard embedding techniques.

\section{Related Work}

There are two general approaches to visualizing temporal data~\cite{Beck2014,Aigner2023}: time-to-space mapping and time-to-time mapping. In time-to-space mapping, time is represented as a spatial element on a static display that uses the color, shape, or size of spatial elements that represent the data points or annotates visualizations with labels or arrows to indicate transitions in time. In contrast, time-to-time mapping represents the time dimension through visual changes of the visualization over time, typically using animations or triggered by user interactions. While this approach may be convenient for interactive applications, it faces challenges related to human-computer interaction and human perception~\cite{Fisher2010}. Here, we focus on time-to-space mappings that result in static visualizations that are easier to interpret and more suitable for dissemination.

There are many ways to visualize time data in a single, static visualization. A classic example are time series visualizations, where time is plotted on the $x$-axis and a dependent variable is plotted on the $y$-axis. While intuitive for univariate time series, this approach becomes less effective with large numbers of variables. To address this, several methods use dimensionality reduction techniques that project high-dimensional time series into a single dimension, which is then stacked along a second time dimension~\cite{Crnovrsanin2009,Jackle2016}. While allowing the entire state of the system to be visualized in a single two-dimensional plot, embedding high-dimensional data sets to a single dimension is often too constrained to adequately reveal complex relations between data points. This concept can be extended to three dimensions, where the temporal evolution of a two-dimensional embedding is plotted along a third, time dimension~\cite{DwyerGallagher2004,Rauber2016,Dadu2022}. Despite their simplicity, such three-dimensional plots have significant limitations that include challenges with viewpoint selection and visual clutter~\cite{Coimbra2016}.

Instead of adding an additional, time dimension, temporal relations can be represented by drawing arrows between consecutive data points in the embedding space, such as in a two-dimensional scatter plot. For example, in a visualization where data points represent individual days of various measurements, their corresponding glyphs can be connected with arrows to illustrate their temporal progression. This approach, commonly known as {\em trace visualization}~\cite{Kondo2014}, is frequently employed in bioinformatics and transcriptomics, where arrows are first averaged and then overlaid on a two-dimensional scatter plot to depict the developmental trajectories of different cell types~\cite{LaManno2018}.

The use of dimensionality reduction techniques to visualize temporal, high-dimensional data is not a new concept. However, little effort has been made to adapt these methods to leverage available temporal information in order to produce visualizations that can more effectively reveal temporal patterns. For example, Elzen \etal~\cite{Elzen2016} applied PCA and t-SNE to visualize dynamic graphs and identified multiple stationary states. Similarly, Ali \etal~\cite{Ali2018} used similar approaches to visualize multivariate time series. Hinterreiter \etal~\cite{Hinterreiter2021} explored the use of t-SNE, UMAP, and ISOMAP to visualize trajectories in tasks such as sorting algorithms, Rubik's Cube solvers, chess games, and neural network training. In transcriptomics, PCA, t-SNE, and UMAP are often used to map the cell landscape and visualize differentiation trajectories~\cite{LaManno2018}. Notably, all of these methods rely on standard implementations of dimensionality reduction algorithms, which do not incorporate available temporal information and may not be optimal for visualizing inherently dynamic data.

In this manuscript, we also propose a set of quantitative criteria for evaluating visualizations of temporal data. The evaluation of dimensionality reduction methods has garnered significant attention in recent years. Numerous approaches have been developed to assess the fidelity of the resulting embeddings, including metrics for distance preservation~\cite{Kruskal1964a,Becht2019}, neighbor rank-based metrics~\cite{VennaKaski2001,Lee2009,Lee2015}, and topological metrics based on persistent homology~\cite{RieckLeitte2015}. 

On the other hand, there are limited guidelines on how to effectively display temporal data in scatter plots, particularly in the context of dimensionality reduction. However, our approach bears significant similarities to directed graph drawing, where entities are positioned on a two-dimensional plane and relationships between them are represented using arrows. Fortunately, the question of what constitutes a good and comprehensible graph visualization has been extensively studied in the graph visualization literature for decades~\cite{Tamassia1988,Ware2002,Purchase2002,TaylorRodgers2005,Bennet2007}. Given these similarities, we rely primarily on metrics from the graph visualization literature to evaluate how effectively our proposed approach produces coherent and informative visualizations.

Several measures have been developed to quantify the legibility of graph drawings. In a survey by Bennet \etal~\cite{Bennet2007}, the authors identified 18 measures, many of which have been empirically validated through user studies, to evaluate the aesthetics and legibility of graphs. These measures include metrics related to vertex placement, such as positioning similar nodes close to each other; edge placement, such as minimizing edge crossings; and overall graph layout, such as maintaining a correct aspect ratio. In this study, we focus solely on measures related to edge placement, as the overall embedding layout and node placements are determined by the t-SNE algorithm.

\section{Methods}

We begin by introducing the relevant notation and providing a brief overview of the t-SNE algorithm. We then present our newly proposed loss functions.

\subsection{t-SNE}

Consider a high-dimensional dataset $\mathbf{X} \in \mathbb{R}^{N \times d}$, where $N$ is the number of data points and $d$ is the dimensionality of each data point. Let $G$ be a directed graph, $G = (V, E)$, where $V$ represents the set of vertices $v_i$, corresponding to individual data points $\mathbf{x}_i$, and $E$ denotes the set of edges $e_{ij}$ representing the temporal connections between data points $i$ and $j$. 

When visualizing high-dimensional datasets, the primary goal is to derive a low-dimensional embedding $\mathbf{Y} \in \mathbb{R}^{N \times 2}$ that preserves the topological features of $\mathbf{X}$. In two-dimensional visualizations, these connections $e_{ij}$ are depicted as directed line segments $\mathbf{p}_{ij}$ (arrows) linking the corresponding data points $i$ and $j$ in the embedding space, where $\mathbf{p}_{ij} = [\mathbf{y}_i, \mathbf{y}_j]$.

$t$-distributed stochastic neighbor embedding (t-SNE)~\cite{Maaten2008} is a non-linear dimensionality reduction technique commonly used to visualize high-dimensional data. t-SNE aims to find a low-dimensional representation $\mathbf{Y}$ such that if two data points are close in the high-dimensional space $\mathbf{X}$, then they are also close in the low-dimensional space $\mathbf{Y}$.

Formally, the t-SNE algorithm seeks a low-dimensional representation $\mathbf{Y^{*}}$ by minimizing the Kullback-Leibler (KL) divergence $D_{\rm KL}(\mathbf{P} \| \mathbf{Q})$ between the pairwise similarities $\mathbf{P}$ in the high-dimensional space $\mathbf{X}$ and the pairwise similarities $\mathbf{Q}$ in the low-dimensional space $\mathbf{Y}$. This optimization problem is expressed as:
\begin{equation}
  \mathbf{Y^{*}} = \arg \min_{\mathbf{Y}} \text{KL}(\mathbf{P} \mid \mid \mathbf{Q}).
  \label{eq:kl_divergence}
\end{equation}
where the goal is to find $\mathbf{Y^{*}}$ that best preserves the similarity structure of $\mathbf{X}$ in the reduced space.

The similarities $\mathbf{P} = [p_{ij}]$ between data points in $\mathbf{X}$ are obtained using the Gaussian kernel,
\begin{equation}
  p_{ij} = \frac{p_{j \mid i} + p_{i \mid j}}{2N}, \quad p_{j \mid i} = \frac{\exp \left ( -\mathcal{D}(\mathbf{x}_i, \mathbf{x}_j ) / 2 \sigma_i^2 \right )}
  {\sum_{k \neq i } \exp \left ( -\mathcal{D}(\mathbf{x}_i, \mathbf{x}_k ) / 2 \sigma_i^2 \right )}, \quad p_{i \mid i} = 0,
  \label{eq:gaussian_kernel}
\end{equation}
where $\mathcal{D}$ is some distance measure and the bandwidth of each Gaussian kernel $\sigma_i$ is selected such that the perplexity $u$ of each conditional distribution $p_{i \mid j}$ matches a user-specified parameter value,
\begin{equation}
  \log \left( u \right) = -\sum_{j} p_{j \mid i} \log \left( p_{j \mid i} \right)
\end{equation}
In the low-dimensional space $\mathbf{Y}$, the similarities $\mathbf{Q} = [q_{ij}]$ are characterized by the $t$-distribution,
\begin{equation}
  q_{ij} = \frac{\left ( 1 + || \mathbf{y}_i - \mathbf{y}_j ||^2 \right )^{-1}}
  {\sum_{k \neq l}\left ( 1 + || \mathbf{y}_k - \mathbf{y}_l ||^2 \right )^{-1}},
  \quad q_{ii} = 0.
  \label{eq:cauchy_kernel}
\end{equation}

\subsection{Directional Losses}

Our aim is to design informative visualizations that facilitate the identification of temporal patterns. The Gestalt principles of human perception~\cite{Koffka1935}, which describe how humans group elements and recognize patterns, provide guidelines for creating visualizations that align with the human perceptual system. Key principles such as proximity, continuity, and common fate guide the grouping of elements and the recognition of patterns. Motivated by these principles, we introduce two loss functions that, when integrated into dimensionality reduction approaches like t-SNE, generate low-dimensional embeddings that are more temporally meaningful.

\subsubsection{Directional Coherence Loss (DCL)}

Before introducing the first of our proposed loss functions, we review two Gestalt principles relevant to visualizing trajectories in two-dimensional embeddings: continuity and similarity. The Gestalt principle of continuity suggests that we are more likely to perceive objects as part of the same unit when they are arranged along a line or curve. Similarly, the principle of similarity indicates that we naturally group objects based on shared visual characteristics, such as color, shape, or orientation. Both principles are directly applicable to two-dimensional temporal visualizations.

First, consider a data point positioned along a trajectory. This point serves as both the starting and ending point for two consecutive trajectory segments, represented by two arrows. When the angle between these arrows is small, they form a nearly straight line, which we perceive as a single, cohesive unit. Conversely, if the angle is large, the arrows create a jagged path, disrupting visual continuity and making it harder to identify the segments as part of the same sequence.

Second, consider a set of arrows within a two-dimensional embedding. When the arrows in a specific region share a similar orientation, we more easily perceive them as forming a single trajectory. With consistent arrow orientation, viewers naturally group these arrows together, interpreting them as part of a coherent trajectory rather than a collection of unrelated trajectories.

These two principles motivate our \textit{Directional Coherence Loss} (DCL)~\cite{Policar2023}, which aims to ensure that arrows close to one another in the low-dimensional embedding space $\mathbf{Y}$ point in approximately the same direction. This promotes the formation of smooth, continuous paths for consecutive arrow segments and encourages nearby segments to align in a single direction, enabling their perceptual grouping into a coherent trajectory. Since each arrow is represented as a line segment parameterized by points $\mathbf{y}_i$ and $\mathbf{y}_j$, directional coherence is achieved by adjusting the positions of these points accordingly.

Let $\mathbf{u}_{ij}$ be the unit vector corresponding to the line segment $\mathbf{p}_{ij} = [ \mathbf{y}_i, \, \mathbf{y}_j ]$,
\begin{equation}
  \mathbf{u}_{ij} = \mathbf{p}_{ij} / ||\mathbf{p}_{ij}||, \quad \mathbf{p}_{ij} = \mathbf{y}_j - \mathbf{y}_i.
\end{equation}
For each pair of edges $e_{ij}$ and $e_{kl}$ in $E$, we can determine the directional coherence of their corresponding arrows in the embedding by computing the dot product $\mathbf{u}_{ij} \cdot \mathbf{u}_{kl} = || \mathbf{u}_{ij} || \: || \mathbf{u}_{kl} || \cos \theta$, where $\theta$ denotes the angle between the two vectors. In our case, $||\mathbf{u}_{ij}|| = ||\mathbf{u}_{kl}|| = 1$, so their dot product simplifies to $\mathbf{u}_{ij} \cdot \mathbf{u}_{kl} = \cos \theta$. When $\mathbf{u}_{ij}$ and $\mathbf{u}_{kl}$ point in the same direction, their dot product is $1$. Conversely, when $\mathbf{u}_{ij}$ and $\mathbf{u}_{kl}$ point in opposite directions, their dot product is $-1$.
Therefore, to achieve good directional coherence for any pair of arrows in $E$, we must maximize the dot product of their corresponding directional vectors.

To make our notion of directional coherence compatible with existing dimensionality reduction approaches, we convert the directional coherence into a strictly positive minimization loss. To convert the maximization into a minimization objective, we multiply the equation with $-1$. To enforce strict-positivity and avoid negative penalties, we add a $+1$ term to the above formulation and shift the domain from $[-1, 1]$ to $[0, 2]$. Additionally, we have found it beneficial to square the resulting quantity, leading to faster convergence and more visually appealing visualizations. The directional coherence loss between edges pair of edges $e_{ij}$ and $e_{kl}$ then becomes
\begin{equation}
  \text{DCL}(\mathbf{p}_{ij}, \mathbf{p}_{kl}) = \left ( - \left ( \mathbf{u}_{ij} \cdot \mathbf{u}_{kl} \right ) + 1 \right ) ^2
  \label{eq:direction}
\end{equation}

Applying the DCL to all pairs of arrows would force all arrows to point in the same direction, regardless of their proximity in the embedding space.
Instead, we want only nearby arrow pairs to point in the same direction, therefore, we penalize only nearby arrow pairs. The distance between two line segments $\mathbf{p}_{ij} = [ \mathbf{y}_i, \, \mathbf{y}_j ]$ and $\mathbf{p}_{kl} = [ \mathbf{y}_k, \, \mathbf{y}_l ]$ is defined as
\begin{equation}
  d(\mathbf{p}_{ij}, \mathbf{p}_{kl}) = \arg \min_{s, t} || \left [ s \cdot \mathbf{y}_i + (1 - s) \cdot \mathbf{y}_j \right ] - \left [ t \cdot \mathbf{y}_k + (1 - t) \cdot \mathbf{y}_l \right ] ||,
\end{equation}
where $s, t \in [0, 1]$. Intuitively, this distance corresponds to the distance between the two closest points on these line segments. If the line segments intersect, then their distance is 0.

We penalize nearby arrow pairs using a Gaussian kernel on the obtained pairwise line-segment distances,
\begin{equation}
 w(\mathbf{p}_{ij}, \mathbf{p}_{kl}) = \frac{1}{\sqrt{2 \pi \sigma^2}} \exp \left ( -d(\mathbf{p}_{ij}, \mathbf{p}_{kl}) / 2 \sigma ^2 \right ),
  \label{eq:weights}
\end{equation}
where $\sigma^2$ is the variance of the Gaussian distribution and is determined via a user-adjustable scale parameter $s$, so that
\begin{equation}
 \sigma^2 = s \cdot \max_{d} \left( \max_i \mathbf{y}_{i,d} - \min_i \mathbf{y}_{i,d} \right).
 \end{equation}
The scale $s$ determines the fraction of the embedding in which we wish arrows to point in the same direction and can greatly affect the final embedding. A large value of $s$ will enforce the DCL across the entire embedding, while small values of $s$ will enforce the DCL more locally, revealing local temporal patterns, and will have a more limited impact on the global layout of the embedding.

Combining the directionality penalty from Eqn.~\ref{eq:direction} and the weights from Eqn.~\ref{eq:weights}, we obtain the final normalized directional coherence loss,
\begin{equation}
 \mathcal{L}_{\text{DCL}} = \frac{1}{\binom{|E|}{2}} \sum_{e_{ij} \in E} \sum_{e_{kl} \in E} w(\mathbf{p}_{ij}, \mathbf{p}_{kl}) \left ( - \left ( \mathbf{u}_{ij} \cdot \mathbf{u}_{kl} \right ) + 1 \right ) ^{2}, \quad (i, j) \neq (k, l).
\end{equation}

\subsubsection{Edge Length Loss (ELL)}

Another important factor that influences the readability of the resulting embeddings is the length of the arrows connecting data points. Edge length has been shown to be an important graph aesthetic, and it is generally agreed upon that shorter arrows produce more readable visualizations~\cite{Purchase2002}. We can further motivate the preference for shorter edges with the Gestalt principle of proximity: minimizing edge lengths encourages connected data points to be positioned closer to one another, allowing them to be perceived as part of a cohesive whole.
To this end, we introduce the \textit{Edge Length Loss} (ELL), which penalizes arrow lengths in the following manner,
\begin{equation}
\mathcal{L}_{\text{ELL}} = \frac{1}{|E|} \sum_{e_{ij} \in E} ||\mathbf{p}_{ij}|| ^{\alpha}.
\label{eq:ell}
\end{equation}
Here, the exponent $\alpha$ is a user-specified length modulation parameter that dictates to what degree longer arrows should be penalized over shorter ones. The effects of this parameter are discussed in more detail in Section~\ref{sec:ell_factor}.

One might wonder why, instead of minimizing edge length, we do not attempt to preserve the edge length from the high-dimensional space. However, this approach would effectively involve preserving distances between specific pairs of data points, which would directly conflict with the t-SNE objective. Instead, we focus on edge length purely as an aesthetic feature, aiming to enhance the overall readability of the resulting visualizations.

\subsection{Optimization}

Both proposed loss functions can be integrated into various dimensionality reduction methods. In this work, we incorporate these losses into the t-SNE algorithm, as implemented in the \textsf{openTSNE}~\cite{Policar2024} Python library. The resulting loss is a combination of the three loss terms,
\begin{equation}
 \mathcal{L} = \mathcal{L}_{\text{t-SNE}} + \lambda \mathcal{L}_{\text{DCL}} + \mu \mathcal{L}_{\text{ELL}},
\end{equation}
where $\lambda$ and $\mu$ denote the penalty strengths for the DCL and ELL.

We optimize each embedding using the same procedure as the t-SNE algorithm. The t-SNE optimization process consists of two phases. In the first phase, known as \textit{early exaggeration}, the attractive forces between data points are amplified by a factor $\rho$, typically set to $12$~\cite{Maaten2008}. In the second, standard phase, the attractive forces are returned to their original values with $\rho=1$. Optimization is carried out using gradient descent with the delta-bar-delta update rule~\cite{Jacobs1988} and a learning rate of $N/\rho$, as proposed by Belkina \etal~\cite{Belkina2019}.

We apply DCL and ELL at each iteration of the optimization process using their respective learning rates. Gradient clipping is applied individually to the gradients of both DCL and ELL to ensure stability during optimization, as the absence of gradient clipping results in unstable behavior. We also increase the number of total iterations in the standard phase of the optimization to 1,500 as the combination of losses is often more difficult to optimize. The remaining settings are left unchanged.

\section{Results and Discussion}

In this section, we introduce appropriate quantitative metrics for evaluating embeddings of time-dependent data, examine the effects of its parameters, and provide guidelines for selecting optimal parameter values.

\subsection{Quantitative Assessment Metrics}

To quantify the quality of our resulting embeddings and compare it with alternative approaches, the evaluation must adopt a two-faceted approach and evaluate both the preservation of data structure and temporal information. 

To evaluate the fidelity of the resulting embeddings with respect to the overall placement of data points, we report the neighborhood-preserving area under the curve (AUC)~\cite{Lee2015}, distance correlation~\cite{Becht2019}, and Spearman distance correlation~\cite{Kruskal1964b} (also known as the Shepard goodness score). These metrics offer insights into both distance preservation and neighbor-ranking capabilities and are well-established tools in the literature for assessing the quality of dimensionality reduction methods. While the AUC provides the most complete assessment of neighborhood preservation, the Pearson correlation score reveals how well actual distances are preserved, while the Spearman correlation reveals the ranking preservation of distances. We primarily rely on the AUC in our assessment, but we also include the Pearson and Spearman correlation scores as they can reveal insights into the behavior of our embedding approach.

To evaluate the temporal coherence of the resulting visualizations, we follow the core ideas for the assessment of graph visualizations, as reviewed in the Related Work. In particular, we consider the following four measures: 
\begin{enumerate}
  \item \textbf{Minimize the number of edge crossings.} The number of edge crossings has been consistently found to be one of the most important indicators of graph readability~\cite{Tamassia1988,Ware2002,Purchase2002}. To quantify this, we report on the number of edge crossings in the embedding.
  \item \textbf{Minimize edge length.} Shorter edge lengths tend to produce more uniform edge lengths, which often induce regular structures, making it easier to group together related nodes. To quantify this, we report the value of the ELL.
  \item \textbf{Minimize continuation angles.} Ware \etal~\cite{Ware2002} found that edge continuity is one of the most important factors in graph path-finding tasks, perhaps even more important than the number of crossings.
 Edge continuity states that angles between consecutive arrow segments should be minimized so that arrows follow smooth, easy-to-track lines. Here, we report the mean angle between two consecutive arrows.
  \item \textbf{Maximize consistent flow direction.} To facilitate flow tracing in directed graphs, arrows should point in the same direction. While this is typically applied globally in graph visualization, we enforce this metric locally, encouraging arrows within each region to align in similar directions. To quantify this, we report the value of the DCL.
\end{enumerate}
We collectively refer to these four metrics as metrics of temporal coherence, as they quantify key aspects that ensure temporal patterns are both legible and discernible in the resulting visualizations.

\subsection{Datasets}

\begin{figure}[thp!]
 \centering
 \includegraphics{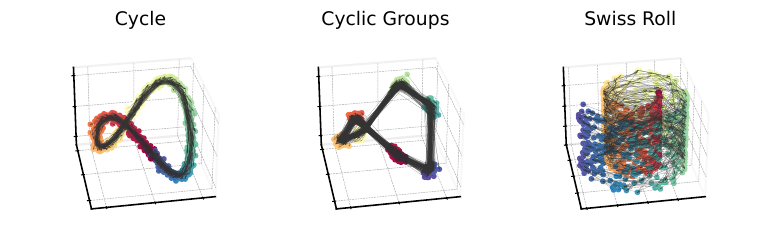}
 \caption{\label{fig:synthetic_data_sets}{\bf The three-dimensional synthetic data sets used throughout the evaluation.}}
\end{figure}

We evaluate our approach using three synthetic datasets depicted in Fig.~\ref{fig:synthetic_data_sets} and two real-world datasets. 

The synthetic datasets include two cyclic datasets and the Swiss Roll dataset, each containing 1,000 samples. Each synthetic data set contains an inherently two-dimensional manifold embedded in three dimensions. The first cyclic dataset consists of points sampled uniformly from a continuous cyclic manifold, with arrows connecting nearby points along the circular trajectory in a single direction. The second cyclic dataset is sampled from the same manifold but consists of six discrete clusters of points, with arrows connecting consecutive clusters. The Swiss Roll dataset, commonly used in dimensionality reduction studies, represents an intrinsically two-dimensional plane curled up and embedded in three-dimensional space. We augment this dataset with arrows connecting data points along a specified dimension of the intrinsic two-dimensional plane. Each synthetic dataset features a relatively simple intrinsic structure but is embedded in three dimensions.

The two real-world datasets include the Columbia Object Image Library (COIL-20) dataset and the COVID-19 dataset from our previous work~\cite{Policar2023}. The COIL-20 dataset consists of 1,440 black-and-white images of twenty objects, each rotated around its axis at 72 distinct viewing angles, with arrows connecting consecutive viewing angles for each object. The COVID-19 dataset is derived from three time series representing daily tests, confirmed cases, and hospitalizations in Slovenia from March 2020 to March 2022. Using the sliding window approach from Ali \etal~\cite{Ali2018}, with a window size and stride of 7, this multivariate time series is converted into a matrix format, resulting in 160 samples, each representing one week of the COVID-19 pandemic in Slovenia.

\subsection{\label{sec:individual_parameters}
The Impact of Individual Parameters}

Incorporating the two directional loss terms into existing dimensionality reduction algorithms introduces four new user parameters, each of which can significantly influence the resulting embeddings. The DCL introduces two parameters: $\lambda$, which controls the strength of the directional coherence enforcement, and $\sigma$, which determines the scale at which it is applied. Similarly, the ELL introduces two parameters: $\mu$, which determines the strength of the arrow length penalty, and $\alpha$, which modulates the penalty's effect. Below, we explore the quantitative and qualitative impacts of each of these parameters.

\subsubsection{\label{sec:dcl_strength}
The Effects of DCL Strength $\lambda$}

\begin{figure}[thp!]
 \centering
 \includegraphics{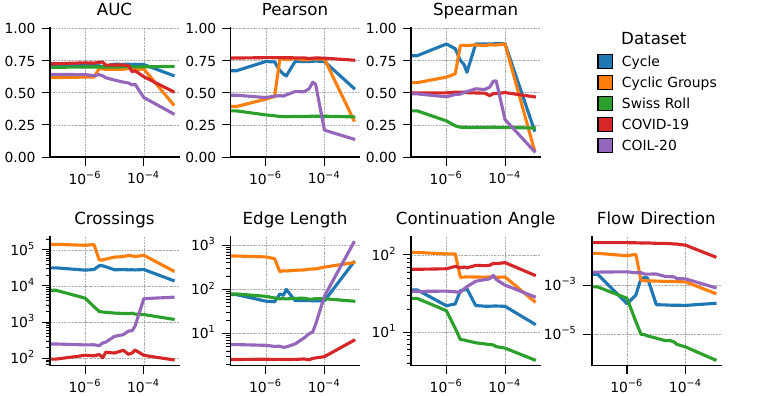}
 \caption{\label{fig:dcl_strength_metrics}{\bf The effects of different DCL strengths $\lambda$ on embedding quality.} The metrics in the top row relate to the fidelity of the embedding, while the metrics in the bottom row quantify the visual temporal coherence of the resulting visualizations.}
\end{figure}

The top row of Fig.~\ref{fig:dcl_strength_metrics} shows the impact of different DCL strengths $\lambda$ on metrics related to the fidelity of the embedding.
While most curves display a slight downward trend for $\lambda > 10^{-3}$ (apart from the Orange curve corresponding to the Cyclic Groups data set, which we discuss later), they appear relatively flat for lower values of $\lambda$. This indicates that enforcing the DCL at lower strengths does not significantly degrade the fidelity of the embedding and disrupts the spatial relationships between the data points.

The bottom row in Fig.~\ref{fig:dcl_strength_metrics} shows how increasing the strength of the DCL penalty affects the metrics related to temporal coherence.
Unsurprisingly, the Flow Direction metric, which is directly minimized by the DCL, improves with higher penalty strengths. Similarly, the Continuation Angle metric, also directly minimized by the DCL, also decreases as the penalty strength increases.
However, the number of crossings, which has been shown to be one of the most important factors in visualization legibility, also tends to decrease in the three synthetic data sets and remains reasonable at intermediate penalty strengths.
For smaller DCL strengths, the Edge Length is not greatly affected but markedly increases on some data sets for higher values of $\lambda$.
Overall, these findings suggest that moderate values of $\lambda$, ranging between $10^{-6}$ and $10^{-4}$, achieve an effective balance between embedding fidelity and temporal coherence.

\begin{figure}[thp!]
 \centering
 \includegraphics{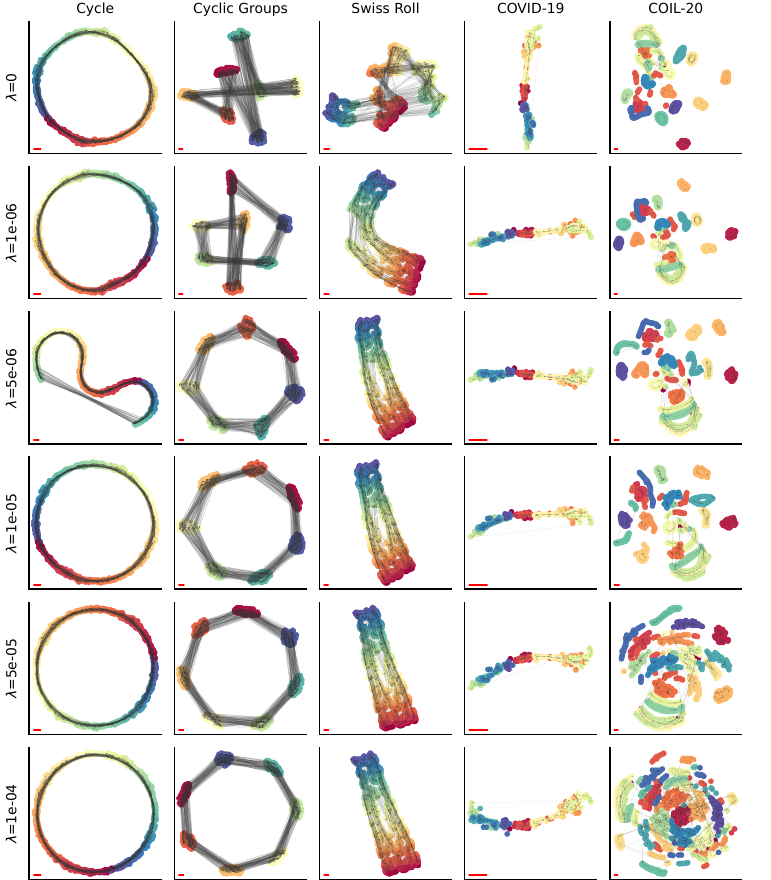}
 \caption{\label{fig:dcl_strength}{\bf The effects of different DCL strengths $\lambda$ on resulting visualizations.} The top row shows a standard t-SNE embedding for reference, while the rows below show the effects of increasing $\lambda$. Point colors indicate temporal progression, except in the COIL-20 data set, in which colors correspond to the 20 different classes. The scale of the embedding is indicated with the red line in the bottom left of each panel. As we increase the DCL strength $\lambda$, many inherently synthetic, two-dimensional manifolds become more apparent. For instance, the Cyclic Groups and Swiss Roll data sets suddenly unravel, and the underlying structure is uncovered. However, using larger values of $\lambda$ can overpower the t-SNE loss and lead to non-sensical visualizations, e.g., the bottom panel of the COIL-20 data set.}
\end{figure}

Fig.~\ref{fig:dcl_strength} shows the effects of increasing the strength of the DCL on the resulting visualizations.
As we increase $\lambda$, data points follow ever smoother trajectories
In synthetic data sets, where the temporal structure follows simple and predictable patterns, the original temporal pattern is recovered even with low penalty strengths.
For instance, in the Cyclic Groups data set, the cyclic pattern is only recovered when setting $\lambda = 5 \cdot 10^{-6}$, which leads to a sharp increase in the Pearson and Spearman metrics, as shown in the top panels of Fig.~\ref{fig:dcl_strength_metrics}.
While the three synthetic data sets remain unchanged for higher penalty values $\lambda$, in the two real-world data sets, the DCL quickly overpowers the t-SNE loss function at high values of $\lambda$ and degrades the overall embedding fidelity, completely obfuscating the spatial relationships between data points.
For example, in the COIL-20 data set, we expect the visualization to reveal 20 distinct rings corresponding to each of the twenty rotating objects. However, as we increase $\lambda$ to $5 \cdot 10^{-5}$, the DCL forces the points to follow circular paths around the center of the embedding. This effect is even more pronounced when increasing $\lambda$ to $10^{-4}$ in which the distinct rings disappear altogether.

\subsubsection{\label{sec:dcl_scale}The Effects of DCL Scale $s$}

A key consideration when applying the DCL is specifying the scale at which it is enforced.
In our prior work~\cite{Policar2023}, we used a fixed kernel bandwidth $\sigma^2$ for the DCL throughout the optimization. However, this approach fails to account for the varying scales that embeddings typically undergo throughout optimization. For instance, a typical t-SNE embedding is initialized with variance $10^{-4}$. However, these embeddings expand significantly during optimization, typically to scales between 10 and 100. Under a fixed bandwidth configuration, the DCL would initially be equally applied to all edges during the early stages of optimization. As the embedding expands, the DCL is applied ever more locally until its effect on the global directional coherence is negligible and affects only intersecting arrows.
To address this issue, we instead use an adaptive scale parameter $s$, which adjusts the kernel bandwidth $\sigma^2$ throughout the optimization. This approach accommodates the varying scales of the embedding throughout the optimization process.

\begin{figure}[thp!]
  \centering
  \includegraphics{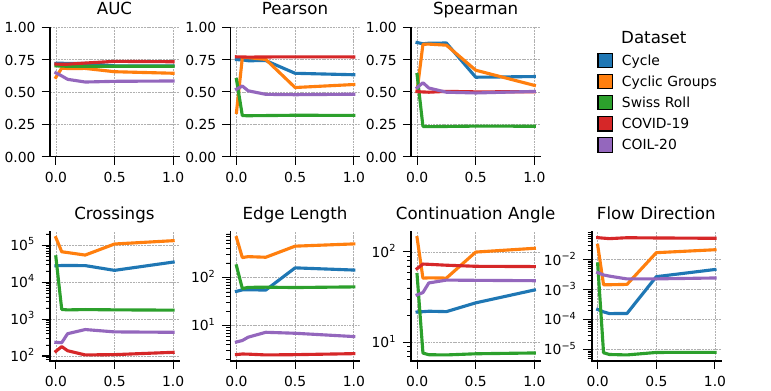}
  \caption{\label{fig:dcl_scale_metrics}{\bf The effects of different DCL scales $s$ on embedding quality.}}
\end{figure}

\begin{figure}[th!]
  \centering
  \includegraphics[width=\linewidth]{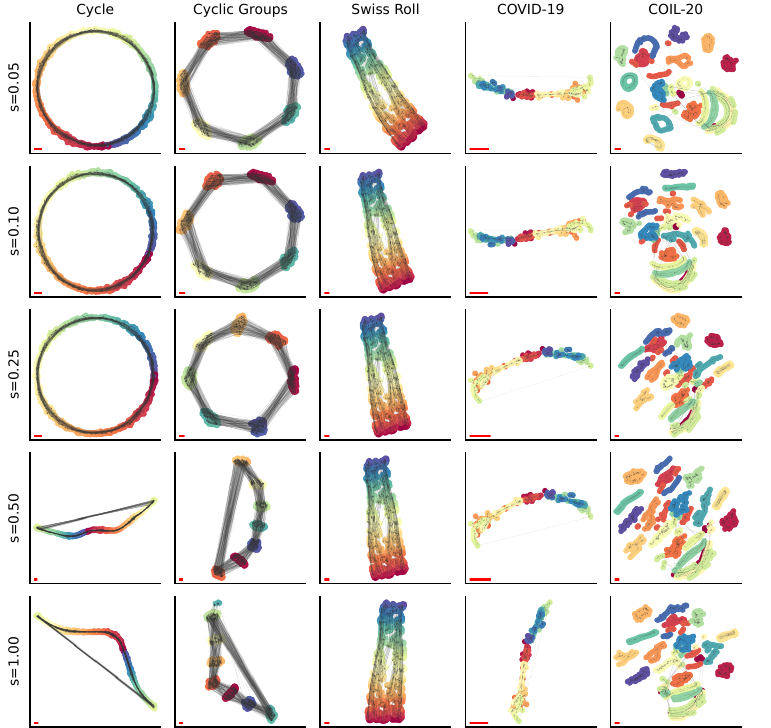}
  \caption{\label{fig:dcl_scale} {\bf The effects of different DCL scale parameter $s$ on resulting visualizations.} Smaller values of $s$ promote local temporal coherence, while larger values of $s$ enforce a consistent directionality of all arrows across the entire embedding.}
\end{figure}

Enforcing the DCL at different scales can serve to highlight different aspects of the temporal progression within the embedding.
At smaller scales, local temporal patterns are highlighted as the DCL is only applied in small regions around each arrow. Conversely, enforcing the DCL at larger scales will result in more globally temporally coherent layouts, emphasizing the temporal progression across the entire embedding.
Figs.~\ref{fig:dcl_scale_metrics} and \ref{fig:dcl_scale} illustrate the effects of varying the scale parameter $s$ on the resulting embeddings. To isolate the effects of the scale parameter, we fix the remaining parameters, setting $\lambda = 2 \times 10^{-5}$ and disabling the ELL loss, setting $\mu = 0$.


For instance, in the Cycle and Cyclic Groups data sets in Fig.~\ref{fig:dcl_scale}, increasing $s$ to $0.5$ and $1$ distorts the cyclic structure and instead produces visualizations in which the trajectories are forced into a more single-direction line shape.
We can observe a similar effect in the COIL-20 data set, where the original circular clusters are deformed into ever more uniform, straight lines, and the cyclic nature of each cluster becomes obscured.
One might expect that such evident distortions would lead to a marked degradation of the embedding fidelity.
Interestingly, however, despite the apparent distortions, the metrics relating to embedding fidelity in Fig.~\ref{fig:dcl_scale_metrics} remain fairly stable at practically all scales $s$.
For data sets that naturally exhibit a single temporal flow, i.e., the Swiss Roll data set and the COVID-19 data set, the overall layout is left largely unchanged even for large values of $s$.

\subsubsection{\label{sec:ell_strength}The Effects of ELL Strength $\mu$}

\begin{figure}[thp!]
 \centering
 \includegraphics{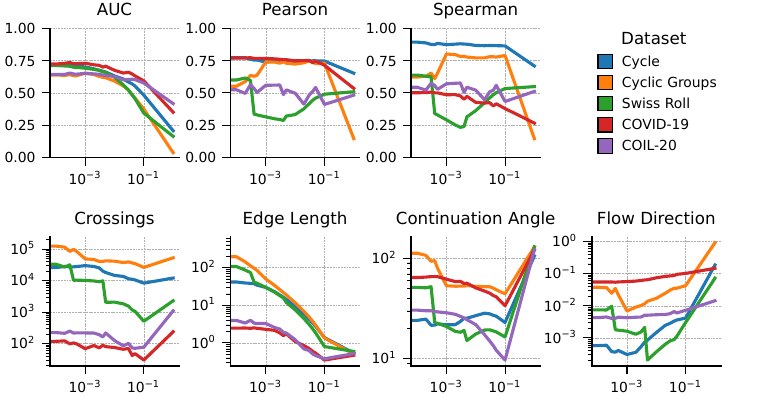}
 \caption{\label{fig:ell_strength_metrics}{\bf The effects of different ELL strengths $\mu$ on embedding quality.}}
\end{figure}

\begin{figure}[thp!]
 \centering
 \includegraphics{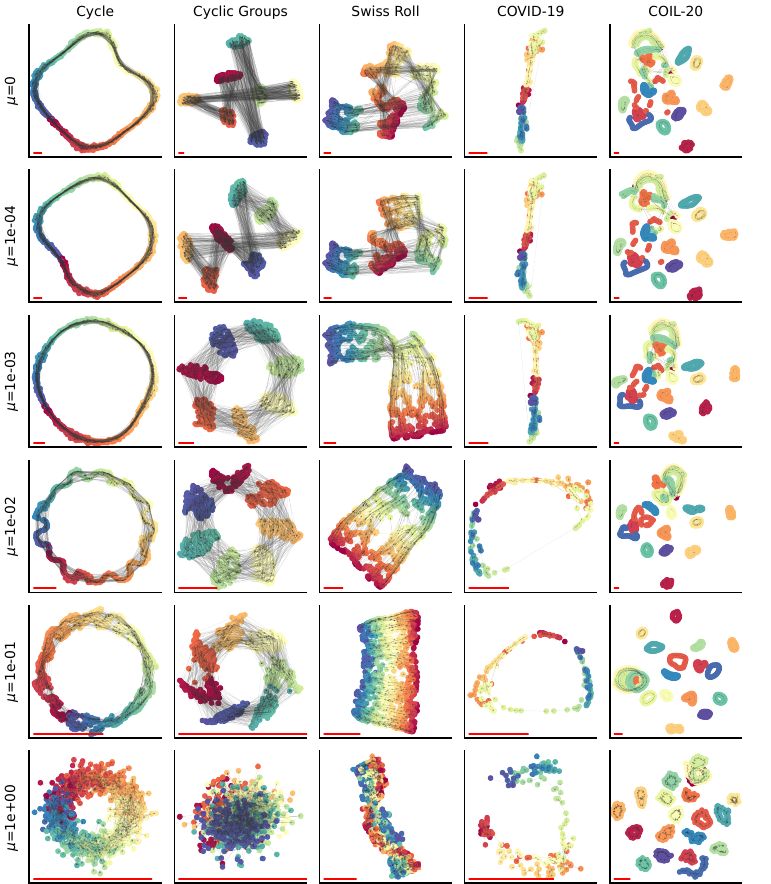}
 \caption{\label{fig:ell_strength}{\bf The effects of different ELL strengths $\mu$ on resulting visualizations.} The top row shows a standard t-SNE embedding for reference while the rows below show the effects of increasing $\mu$. As we increase the ELL strength $\mu$, the arrows connecting data points progressively shrink. At moderate values, the resulting visualizations exhibit smoother trajectories}
\end{figure}

Fig.~\ref{fig:ell_strength_metrics} demonstrates how different ELL strengths $\mu$ affect the embedding quality.
Unlike the DCL, which does not directly affect distances between data points, the ELL explicitly forces connected data points to be positioned closer to one another. This interference in the distances of data points in the embedding likely conflicts more strongly with the t-SNE objective than the DCL.
As such, it is reasonable to assume that its effect on the embedding fidelity is more severe.
The top panels in Fig.~\ref{fig:ell_strength_metrics} confirm this suspicion as the metrics relating to the embedding fidelity exhibit consistent decreases as the ELL strength $\mu$ increases in the COVID-19, COIL-20, and Cycle data sets.
This does not hold for the Cyclic Groups and Swiss Roll data sets. An inspection of Fig.~\ref{fig:ell_strength} provides insight into why this might be the case. At lower penalty values $\mu$, the overall layout of the embedding largely follows the layout produced by t-SNE alone. By itself, the t-SNE objective is unable to untangle the underlying manifold for these particular data sets. Due to the temporal structure present in these data sets, enforcing the ELL more strongly has the added benefit of untangling these manifolds and, as such, results in a better overall embedding structure, which is reflected in the fidelity metrics in Fig.~\ref{fig:ell_strength_metrics}.

The ELL strength also favorably affects the temporal coherence metrics shown in the bottom row of Fig.~\ref{fig:ell_strength_metrics}. For most data sets, the number of crossings and the continuation angles decrease as we increase $\mu$ to about $10^{-1}$. As expected, the edge length metric also decreases, as it is the quantity optimized by the ELL. Interstingly, however, increasing the ELL strength adversely affects the Flow Direction metric, indicating that the two address different yet complementary aspects of temporal coherence.

As we increase the ELL penalty strength $\mu$, the distances between data points decrease, reducing the overall scale of the embedding. This is visible from the red scale bars at the bottom of each panel in Fig.~\ref{fig:ell_strength}. At extreme values, any visible cluster structure of the data points can be severely distorted. For instance, in the bottom row of the Cyclic Groups data set, any semblance of clusters is entirely absent, and the visualization resembles random noise, underscoring the potential trade-off between enforcing ELL and preserving meaningful spatial relationships.

\subsubsection{\label{sec:ell_factor}The Effects of ELL Modulation $\alpha$}

We next examine the role of the ELL modulation parameter $\alpha$, which controls the degree and manner in which the lengths of arrows are penalized. For smaller values of $0 < \alpha < 1$, longer arrows are penalized less severely than shorter arrows relative to their lengths, allowing the formation of longer arrows that connect different regions of the embedding space. On the other hand, larger values of $1 < \alpha$ penalize longer arrows more severely, and shorter arrows are preferred. This prevents the formation of longer arrows and forces connected data points to be positioned more closely to one another, resulting in a smoother and more connected trajectory.

\begin{figure}[th!]
 \centering
 \includegraphics{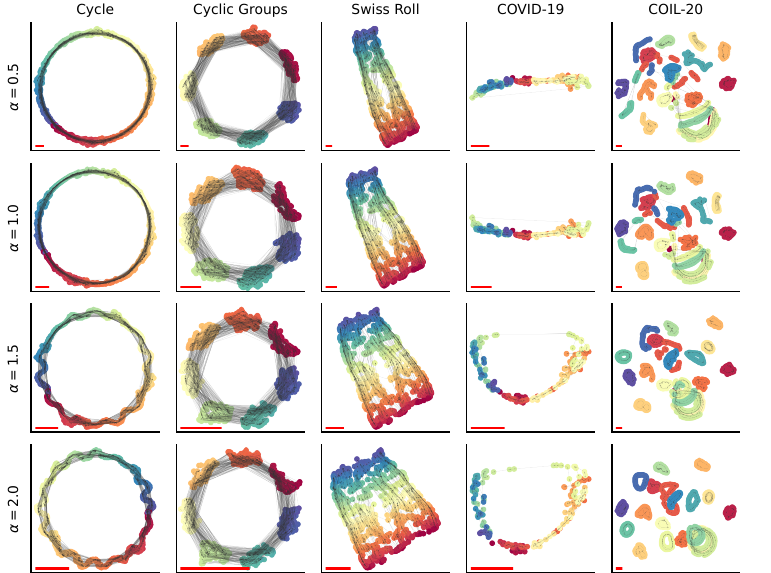}
 \caption{\label{fig:ell_factor}{\bf The effect of increasing the ELL modulation factor $\alpha$.} Lower values of $\alpha$ are more permissive to longer arrows which can connect different regions of the embedding space, while larger $\alpha$ factors penalize the formation of longer arrows and result in a more evenly spaced temporal trajectory. Additionally, higher values of $\alpha$ result in points being positioned closer to one another, resulting in more compact embeddings, as indicated by the red scale bars.}
\end{figure}

Fig.~\ref{fig:ell_factor} illustrates the effects of different ELL parameter values $\alpha$ on the resulting visualizations.
To isolate the effects of $\alpha$, we hold the remaining parameters constant, setting $\mu = 10^{-2}$, $\lambda = 10^{-5}$, and $\sigma = 0.1$.
Note that we enforce the DCL penalty so that the general layout of the Cyclic Groups and Swiss Roll data sets are recovered at all ELL factor values so the evaluation metric values do not reflect the sudden recovery of the underlying temporal patterns.

\begin{figure}[thp!]
 \centering
 \includegraphics{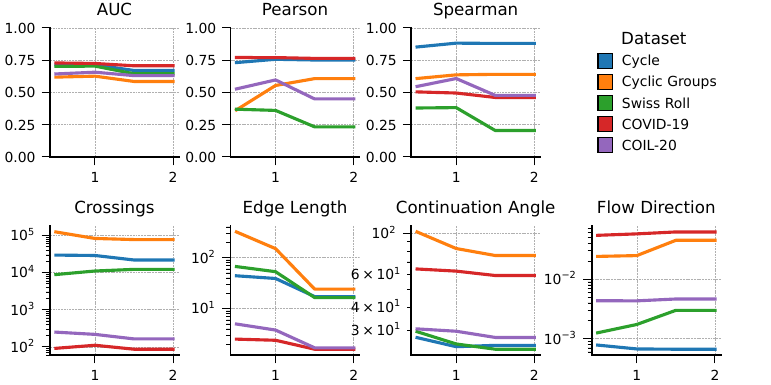}
 \caption{\label{fig:ell_factor_metrics}{\bf The effects of different ELL modulation factors $\alpha$ on embedding quality.}}
\end{figure}

Lower values of $\alpha$ distribute errors across different arrow lengths more uniformly, allowing the formation of longer arrows. As we increase $\alpha$, we progressively increase the penalization of longer arrows, resulting in visualizations where connected points are pulled closer together.
This behavior is best exemplified in the COVID-19 example in Fig.~\ref{fig:ell_factor}. At a low modulation factor $\alpha = 0.5$, the data points follow a slightly curved line from one corner of the embedding to the other, with a single long arrow connecting the two tips of the line. As we increase $\alpha$, this connecting arrow becomes shorter and shorter, pulling the two tips together, forming an arc for $\alpha = 1.5$ and a near circle for $\alpha = 2$.

Similarly to what we observed when increasing the ELL penalty strength $\mu$, a side effect of penalizing longer arrows is that the resulting embeddings become more compact. As longer arrows are more strongly penalized, on average, arrows become shorter, and points are positioned ever closer to one another, reducing the overall scale of the embedding.
This effect is most apparent in the Cyclic Groups example in Fig.~\ref{fig:ell_factor} where increasing $\alpha$ has minimal effect on the overall layout but increasingly compresses the embedding, as indicated by the red scale bars. Forcing connected data points to lie close to one another is sometimes likely in direct conflict with the t-SNE loss function, which enforces a minimum distance between data points. Quantitatively, this results in the sharp deterioration of the overall embedding fidelity, as indicated by the decreasing AUC and Spearman coefficient scores.
This effect is less pronounced in the remaining data sets, where all arrow lengths are about the same length by design, and increasing the modulation factor will affect all arrows equally, resulting in minimal changes to the layout or embedding quality.

\subsection{What happens when we don't have any meaningful temporal structure\label{sec:shuffled_parameters}}

In our experiments thus far, we have only considered data sets that exhibit well-defined, meaningful temporal structures. However, some real-world data sets may lack clear, well-defined temporal structure. In these cases, it is vital that our approach does not produce misleading visualizations, revealing artificial temporal and structural patterns where there are none.

To evaluate the robustness of our approach and identify reasonable parameter values for $\lambda$ and $\mu$, we consider the same five data sets as before but randomly assign arrow endpoints. In this artificially extreme example, the structural patterns inherent in the data set are entirely uncorrelated from the temporal ordering of the data points. While such scenarios are unlikely in practice, as we would typically expect some degree of time dependence in real-world data, this test proves insightful in evaluating the robustness of our approach.

\begin{figure}[thp!]
 \centering
 \includegraphics{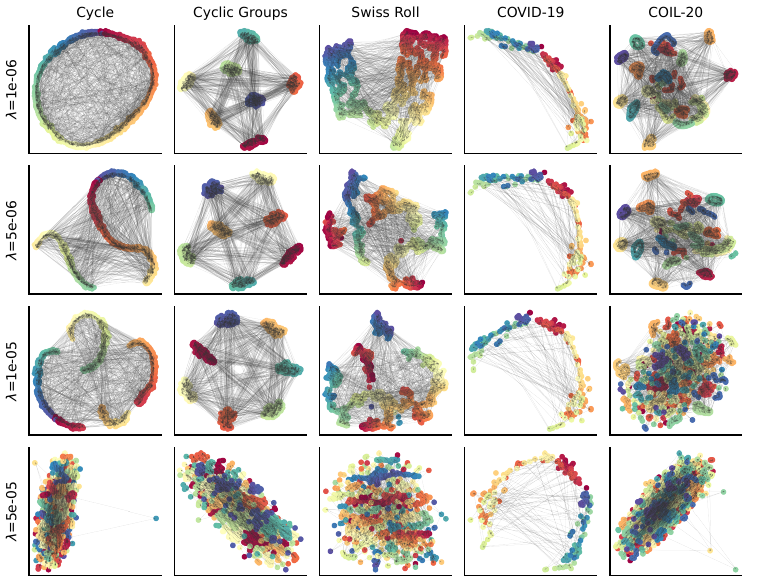}
 \caption{\label{fig:shuffled_dcl_strength}{\bf The effects of different DCL strengths $\lambda$ on resulting visualizations when the data sets contain no temporal patterns.}}
\end{figure}

\begin{figure}[thp!]
 \centering
 \includegraphics{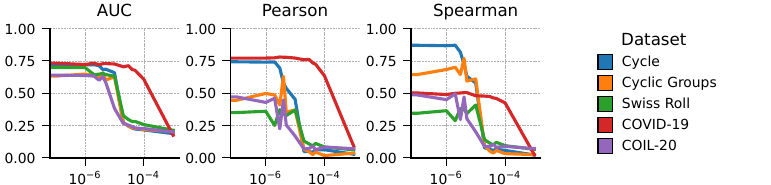}
 \caption{\label{fig:shuffled_dcl_strength_metrics}{\bf The effects of different DCL strengths $\lambda$ on embedding quality when the data sets contain no temporal patterns.}}
\end{figure}

\begin{figure}[thp!]
 \centering
 \includegraphics{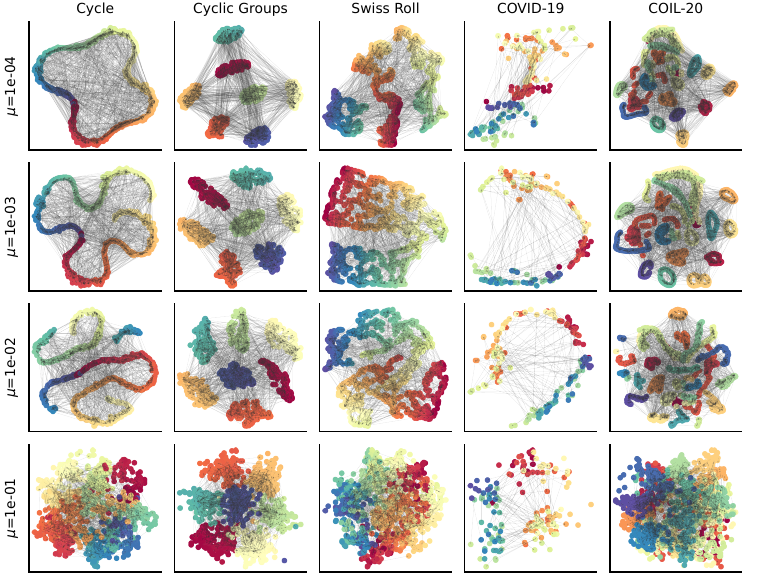}
 \caption{\label{fig:shuffled_ell_strength}{\bf The effects of different ELL strengths $\mu$ on resulting visualizations when the data sets contain no temporal patterns.}}
\end{figure}

\begin{figure}[thp!]
 \centering
 \includegraphics{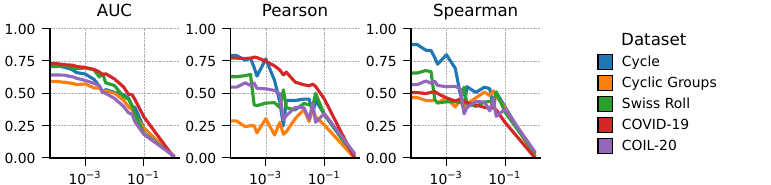}
 \caption{\label{fig:shuffled_ell_strength_metrics}{\bf The effects of different ELL strengths $\mu$ on embedding quality when the data sets contain no temporal patterns.}}
\end{figure}

Figs.~\ref{fig:shuffled_dcl_strength} and \ref{fig:shuffled_ell_strength} show visualizations of the randomized datasets at varying levels of DCL and ELL strengths, while Figs.~\ref{fig:shuffled_dcl_strength_metrics} and \ref{fig:shuffled_ell_strength_metrics} assess their impacts on the quantitative metrics.


Fig.~\ref{fig:shuffled_dcl_strength} shows that even at relatively moderate DCL strengths $\lambda$, the overall topology of the embedding can sometimes become distorted. For instance, standard t-SNE is able to recover the circular topology of the Cycle dataset without the addition of the DCL or ELL losses, as observed in the top panel of Fig.~\ref{fig:ell_strength}. However, when applying a relatively moderate DCL strength of $\lambda=5\cdot10^{-6}$, this circular structure is no longer preserved.
Interestingly, despite this apparent degradation in the global structure, the fidelity metrics in Fig.~\ref{fig:shuffled_dcl_strength_metrics} indicate that the overall embedding fidelity is still largely preserved. All three fidelity metrics remain relatively stable at these strengths,  showing little change until $\lambda=10^{-5}$, at which point the embedding fidelity quickly deteriorates. We can confirm this qualitatively by inspecting the lower panels in Fig.~\ref{fig:shuffled_dcl_strength} in which the produced visualizations resemble a random arrangement of data points.


While enforcing the DCL resulted in a noticeable degradation of the overall global topology of the resulting embeddings, enforcing the ELL on data sets with random temporal patterns appears less detrimental. For instance, the overall topology of all five data sets is largely preserved for moderate strengths $\mu=10^{-4}$ and $\mu=10^{-3}$ in Fig.~\ref{fig:shuffled_ell_strength}. However, for larger values of $\mu$, embedding fidelity deteriorates significantly, and the visualizations lose any recognizable structure.

Inspecting the fidelity metrics in Fig.~\ref{fig:shuffled_ell_strength_metrics} shows that, similarly as observed in Section~\ref{sec:individual_parameters}, enforcing the ELL leads to a gradual but steady decrease in embedding fidelity while the DCL exhibits a harsher transition, dropping significantly at a critical value of $\lambda$.

\subsection{Guidelines to Selecting Appropriate Parameter Values}

Based on the analysis above, we offer the following set of guidelines when selecting parameter values for the DCL and ELL losses:
\begin{enumerate}
    \item Recommended DCL Strength:
    \begin{itemize}
        \item[--] Set $\lambda$ within the range of $10^{-6}$ to $10^{-5}$ for most datasets.
        \item[--] Use higher values only if strong temporal correlations are evident.
    \end{itemize}

    \item Recommended DCL Scale:
    \begin{itemize}
        \item[--] Use $s=0.05$ (5\% of the embedding span) for the DCL scale parameter to emphasize local temporal patterns.
        \item[--] In practical applications, enforcing a single directionality yields little insight. Prefer a local application via lower scale values $s$ to reveal local patterns.
    \end{itemize}

    \item Recommended ELL Strength:
    \begin{itemize}
        \item[--] Set $\mu$ within the range of $10^{-4}$ to $10^{-3}$ for most data sets.
        \item[--] Apply larger values cautiously, especially for datasets without temporal structure.
    \end{itemize}

    \item Default ELL Factor:
    \begin{itemize}
        \item[--] Use $\alpha=1$ for forming longer arrows that connect different embedding regions while preserving proximity.
        \item[--] Use $\alpha=1.5$ for uniform arrow lengths, suitable for smooth trajectory visualization.
    \end{itemize}

    \item Optimization:
    \begin{itemize}
        \item[--] Apply gradient clipping to both DCL and ELL gradients, setting the maximum step size to 1.
        \item[--] Run optimization for more iterations than for the standard method. For t-SNE, we have found that 1,500 iterations usually suffices.
    \end{itemize}

    \item Consider Temporal Correlation:
    \begin{itemize}
        \item[--] Apply this approach only to data sets in which meaningful temporal structure is assumed.
        \item[--] Compare the resulting embeddings to those produced by standard t-SNE and ensure the global spatial relationships, e.g. clusters, are not distorted.
    \end{itemize}
\end{enumerate}

\subsection{Comparison to Existing Dimensionality Reduction Methods}

Several existing publications investigate the usage of dimensionality reduction approaches for visualizing temporal data in a similar manner to the approach proposed here. However, these typically use off-the-shelf implementations of dimensionality reduction approaches, which are not tailored to the dynamic nature of the data.
For instance, Elzen \etal~\cite{Elzen2016} visualize the changing nature of dynamic graphs with PCA and t-SNE while Ali \etal~\cite{Ali2018} use PCA to visualize multivariate time series. Hinterreiter \etal~\cite{Hinterreiter2021} visualize game-move trajectories with t-SNE, UMAP, and ISOMAP.
In the bioinformatics literature, t-SNE and UMAP are the methods of choice when visualizing gene-expression data.

\begin{figure}[thp!]
  \centering
  \includegraphics{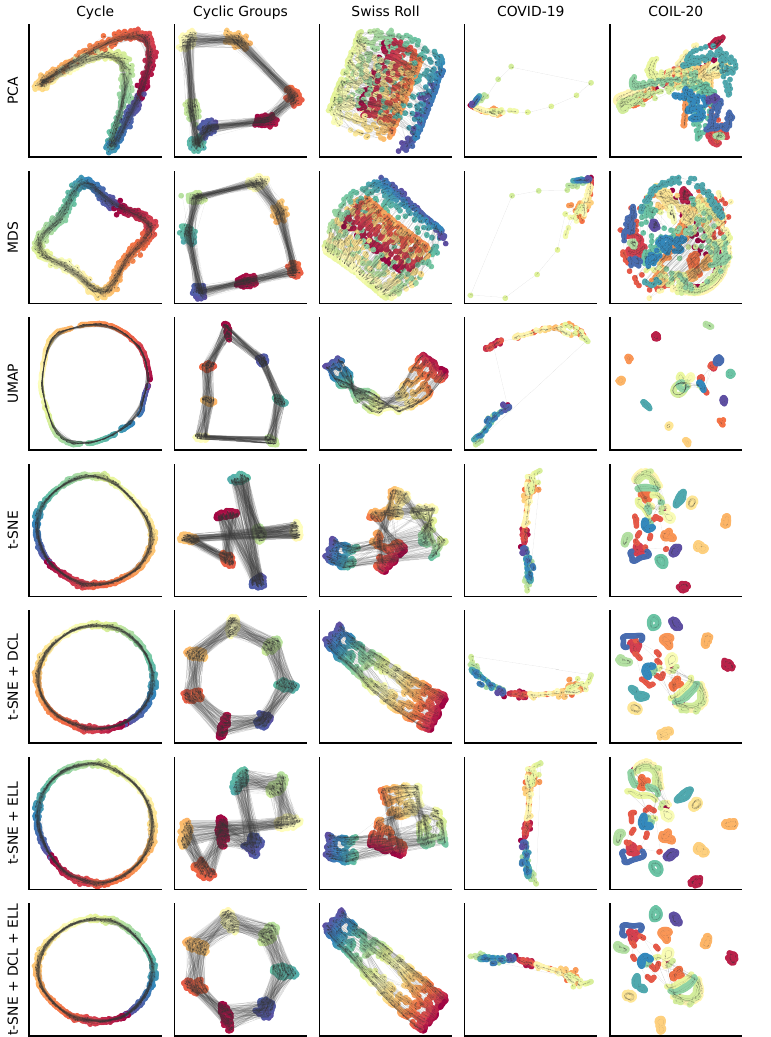}
  \caption{\label{fig:method_comparison}{\bf The visualizations produced by other, popular dimensionality reduction approaches compared to our proposed approaches.}}
\end{figure}

In Fig.~\ref{fig:method_comparison}, we apply several of these popular approaches, including PCA, MDS, UMAP, and t-SNE, to our five different data sets and perform a quantitative evaluation in Table~\ref{tab:method_comparison}. To obtain UMAP embeddings, we use 15 neighbors and a minimum distance of $0.1$. We initialize the embedding via spectral embedding and run optimization for 500 iterations. For the t-SNE embeddings, we use the standard perplexity of 30, initialize the embedding randomly, and run the optimization for 1,500 iterations. We apply a DCL strength $\lambda=10^{-6}$ and DCL scale of $s=0.05$. For the ELL, we apply an ELL strength $\mu=10^{-4}$ using a modulation factor $\alpha=1.5$.

The limitations of linear methods, such as PCA, have long been well understood. Due to their linear nature, these methods can only recover linear manifolds in the high-dimensional space, resulting in poor performance on inherently non-linear manifolds, e.g., the Swiss Roll data set and, to some extent, the cyclic data sets.
While MDS alleviates the limitations of linearity, it often performs poorly on many real-world high-dimensional data sets, in this case, the COIL-20 data set.
These challenges have led to the development of new approaches, including various neighbor-embedding methods such as t-SNE and UMAP, which are often able to better capture complex non-linear structures in the data.
However, none of these approaches incorporate the temporal nature of the data into the embedding construction process.
This is reflected in the relatively poor temporal coherence metric scores achieved by these methods.
Incorporating the DCL and ELL loss terms into the t-SNE algorithm markedly improves these metrics.

\begin{table}[tbp]
  \begin{tabular}{rr|rrrrr}
  \hline
  & & Cycle & Cyclic Groups & Swiss Roll & COVID-19 & COIL-20 \\
   \midrule
  \multirow{7}{*}{\rotatebox[origin=c]{90}{AUC}}& PCA & 0.56 & 0.54 & 0.82 & 0.76 & 0.40 \\
  & MDS & 0.69 & 0.58 & 0.85 & 0.79 & 0.42 \\
  & UMAP & 0.63 & 0.58 & 0.27 & 0.61 & 0.46 \\
  & t-SNE & 0.72 & 0.61 & 0.70 & 0.72 & 0.65 \\
  & t-SNE + DCL & 0.72 & 0.68 & 0.70 & 0.73 & 0.64 \\
  & t-SNE + ELL & 0.72 & 0.65 & 0.71 & 0.73 & 0.65 \\
  & t-SNE + DCL + ELL & 0.72 & 0.67 & 0.70 & 0.73 & 0.64 \\
  \midrule
  \multirow{7}{*}{\rotatebox[origin=c]{90}{Pearson}}& PCA & 0.78 & 0.75 & 0.80 & 0.99 & 0.80 \\
  & MDS & 0.81 & 0.85 & 0.84 & 1.00 & 0.82 \\
  & UMAP & 0.74 & 0.69 & 0.30 & 0.64 & 0.16 \\
  & t-SNE & 0.75 & 0.34 & 0.60 & 0.77 & 0.52 \\
  & t-SNE + DCL & 0.74 & 0.76 & 0.32 & 0.77 & 0.48 \\
  & t-SNE + ELL & 0.75 & 0.62 & 0.55 & 0.77 & 0.52 \\
  & t-SNE + DCL + ELL & 0.75 & 0.75 & 0.32 & 0.77 & 0.49 \\
  \midrule
  \multirow{7}{*}{\rotatebox[origin=c]{90}{Spearman}}& PCA & 0.80 & 0.80 & 0.52 & 1.00 & 0.78 \\
  & MDS & 0.91 & 0.92 & 0.53 & 1.00 & 0.81 \\
  & UMAP & 0.88 & 0.81 & 0.59 & 0.38 & 0.22 \\
  & t-SNE & 0.88 & 0.51 & 0.63 & 0.50 & 0.53 \\
  & t-SNE + DCL & 0.88 & 0.86 & 0.23 & 0.50 & 0.50 \\
  & t-SNE + ELL & 0.88 & 0.65 & 0.59 & 0.50 & 0.54 \\
  & t-SNE + DCL + ELL & 0.88 & 0.85 & 0.24 & 0.51 & 0.51 \\
  \midrule
  \multirow{7}{*}{\rotatebox[origin=c]{90}{Num. Crossings}}& PCA & 24,409 & 63,951 & 15,566 & 70 & 1,014 \\
  & MDS & 21,075 & 53,660 & 6,435 & 74 & 3,876 \\
  & UMAP & 36,958 & 46,345 & 19,491 & 146 & 365 \\
  & t-SNE & 27,504 & 158,541 & 48,026 & 134 & 234 \\
  & t-SNE + DCL & 28,108 & 44,121 & 1,852 & 117 & 259 \\
  & t-SNE + ELL & 27,765 & 109,301 & 35,286 & 112 & 226 \\
  & t-SNE + DCL + ELL & 27,183 & 45,064 & 1,835 & 131 & 237 \\
  \midrule
  \multirow{7}{*}{\rotatebox[origin=c]{90}{Edge Length}}& PCA & 0.42 & 0.95 & 1.24 & 139597.30 & 10975.11 \\
  & MDS & 0.36 & 1.14 & 1.35 & 149800.25 & 187880.84 \\
  & UMAP & 8.37 & 39.38 & 7.29 & 1.05 & 0.16 \\
  & t-SNE & 51.29 & 666.60 & 173.46 & 2.50 & 4.56 \\
  & t-SNE + DCL & 53.69 & 249.59 & 58.59 & 2.21 & 5.27 \\
  & t-SNE + ELL & 48.85 & 202.21 & 116.79 & 2.52 & 5.08 \\
  & t-SNE + DCL + ELL & 48.98 & 156.04 & 48.35 & 2.55 & 5.27 \\
  \midrule
  \multirow{7}{*}{\rotatebox[origin=c]{90}{Cont. Angle}}& PCA & 40.67 & 66.02 & 66.42 & 56.97 & 50.37 \\
  & MDS & 33.18 & 52.16 & 70.79 & 59.52 & 68.40 \\
  & UMAP & 21.64 & 54.92 & 25.46 & 79.53 & 45.02 \\
  & t-SNE & 21.64 & 144.04 & 56.30 & 65.38 & 33.48 \\
  & t-SNE + DCL & 22.03 & 51.71 & 7.61 & 67.52 & 31.81 \\
  & t-SNE + ELL & 21.90 & 102.51 & 46.97 & 64.43 & 33.43 \\
  & t-SNE + DCL + ELL & 21.98 & 51.93 & 7.92 & 67.69 & 31.87 \\
  \midrule
  \multirow{7}{*}{\rotatebox[origin=c]{90}{Flow Direction}}& PCA & 0.5655 & 0.4420 & 0.2450 & 0.0001 & 0.0001 \\
  & MDS & 0.0596 & 0.1668 & 0.2225 & 0.0001 & 0.0000 \\
  & UMAP & 0.0009 & 0.0171 & 0.0042 & 0.1014 & 0.0224 \\
  & t-SNE & 0.0002 & 0.0300 & 0.0072 & 0.0539 & 0.0037 \\
  & t-SNE + DCL & 0.0002 & 0.0017 & 0.0000 & 0.0567 & 0.0032 \\
  & t-SNE + ELL & 0.0002 & 0.0263 & 0.0081 & 0.0538 & 0.0036 \\
  & t-SNE + DCL + ELL & 0.0002 & 0.0023 & 0.0000 & 0.0528 & 0.0033 \\
  \bottomrule
  \caption{\label{tab:method_comparison}We evaluate popular dimensionality reduction methods in tersm of embedding fidelity as well as the temporal coherence in the resulting visualizations.}
\end{tabular}
\end{table}

\section{Conclusion}

We have addressed the challenge of identifying temporal patterns in the representation of multivariate data in low-dimensional, nonlinear embeddings. Temporal relationships are often visualized using arrows to indicate state transitions over time. However, such visual elements can clutter data displays and obscure the underlying patterns. Existing dimensionality reduction techniques do not account for the temporal nature of data. To address this gap, we introduce two novel loss terms: the \textit{directional coherence loss} (DCL) and the \textit{edge length loss} (ELL). These loss terms can be seamlessly integrated with existing dimensionality reduction methods, such as t-SNE, to incorporate temporal information directly into the embedding process, resulting in clearer visualizations of temporal patterns.

This work opens several avenues for future research.
First, the two loss terms enforce directional coherence by affecting the positions of the data points in the two-dimensional embedding. While this approach is viable for some data sets, such an arrangement may be difficult to achieve in the presence of more complex temporal patterns.
Secondly, the DCL exhibits quadratic scaling in the number of connections between data points, making it unsuitable for the visualization of large data sets.
Modern implementations of popular nonlinear dimensionality reduction may address data sets containing up to millions of data points~\cite{Policar2024,McInnes2018}.
In its current form, the DCL scales with quadratic time complexity, allowing it to be applied to data sets containing, at most, several thousands of data points. Developing more efficient implementations and approximation schemes would enable the approach outlined in this manuscript to be applied to larger data sets.
Due to the local nature of the DCL, approximation schemes could be developed that would only compute the interaction between nearby line segments.
Thirdly, we only consider straight line segments connecting two data points; however, the Gestalt principle of continuity would be equally, if not more, applicable if curved arrows were used. For instance, the Time Curve paradigm~\cite{Bach2016} suggests using placing samples along continuous curves. Additionally, they suggest that the length of the curve segments between individual data points could be used to indicate the duration between the two timestamped samples. In this way, the user may not only see the temporal progression of trajectories but also discern the time between subsequent events.

\backmatter

\bmhead{Funding}

This work was supported by the Slovenian Research Agency Program Grant P2-0209 and Project Grant V2-2272.

\bmhead{Conflict of Interest}
The authors report no conflict of interest.

\bmhead{Author Contribution}
P.G.P. developed the approach and ran experiments. P.G.P and B.Z. wrote and reviewed the manuscript.

\bibliography{references}

\end{document}